# A prototype Malayalam to Sign Language Automatic Translator

**Jestin Joy**
AI Lab
Department of Computer Applications
CUSAT
jestinjoy@gmail.com

**Kannan Balakrishnan**
Head of the Department
Department of Computer Applications
CUSAT
mullayilkannan@gmail.com

## Abstract

Sign language, which is a medium of communication for deaf people, uses manual communication and body language to convey meaning, as opposed to using sound. This paper presents a prototype Malayalam text to sign language translation system. The proposed system takes Malayalam text as input and generates corresponding Sign Language. Output animation is rendered using a computer generated model. This system will help to disseminate information to the deaf people in public utility places like railways, banks, hospitals etc. This will also act as an educational tool in learning Sign Language.

## 1 Introduction

According to World Federation of the Deaf (WFD), there are approximately 70 million deaf people in the world. Census 2001 estimates no people with hearing disability at 1,261,722. But the number of sign language interpreters is only about 250; or roughly one for every 72,000 people. (Dasgupta and Basu(2008)). Unlike other languages, sign language doesnt have any written form. Current methods need human intervention for converting text to sign language. This is a challenging tas, considering the scarcity of the sign language interpreters, . Information Technology can effectively be used for this purpose. Also another main challenge from Malayalam perspective is the lack of standardisation. Because of this, practitioners in different parts of Kerala use different approaches in sign language. The objective of this paper is to present a prototype automatic Malayalam text to Sign Language machine translation system. The output is rendered as a three dimensional computer model. User can type Malayalam text into the text box given and system automatically translates it to the corresponding sign language.

According to sign language educators in Kerala, people in different parts of Kerala use sign language in different forms. There are no accepted standard for sign language used in Kerala. The signs used are mainly adopted from Indian Sign Language (ISL). Our system could also be used as a standardised tool for learning sign language in Kerala. This paper is the first attempt in the development of a system for conveting Malayalam text to sign language.

This paper is organized as follows. Section II gives a brief introduction about sign language in Kerala. Then we discuss about current reserch in this field. After that we discuss about design of the proposed system. Then we have results and conclusion.

## 2 Sign Language in Kerala

Like any other sign language, sign language practiced in Kerala uses facial signs in parallel with hand signs. Indian Sign Language also follows this approach. Malayalam, which follows subject-object-verb (SOV) structure, follows the same structure as ISL. (Crystal(1998); George(1972)) According to practisioners, Kerala does not have a sign language as such. Most features are derived from ISL. Even people in differen parts of Kerala use sign language in different forms. This calls for the development of a standardised form of sign language for use in Kerala. Our system could be used as a standardised tool for learning sign language.

## 3 Existing Research

Early work on text to sign language depended mainly on pre recorded videos for conversion. (ASLLRP(2014); Neidle(2000)). Our study reveals that, there are no systems for converting Malayalam text to sign language currently. But there are systems for converting text to ISL in other languages.

Just before the animation stage, current systems make use of an intermediate format for storing sign language. HamNoSys(Hamburg Notation System), developed at University of Hamburg is a widely used one. Projects like ViSiCAST use this notation for generating gestures. (Elliott et al.(2000)Elliott, Glauert, Kennaway, and Marshall) Our system follows an approach of directly converting user input to animation, without using any intermiedate notation.

TESSA is a system for translating speech to British Sign Language (BSL). (Cox et al.(2002)Cox, Lincoln, Tryggvason, Nakisa, Wells, Tutt, and Abbott) It is meant for use post office and follows a phrase lookup approach. San Segundo et al.(2006)San Segundo, Barra-Chicote, Luis Fernando, Montero, de Córdoba, and Ferreiros; San-Segundo et al.(2012)San-Segundo, Montero, Crdoba, Sama, Fernndez, DHaro, Lpez-Ludea, Snchez, and Garca; San-Segundo et al.(2008)San-Segundo, Montero, Macías-Guarasa, Córdoba, Ferreiros, and Pardo) proposed a system for converting spanish to sign language using a rule based approach. Most of the systems developed are domain specific.

INdian Gestural Interaction Translator (INGIT) is a system for converting Hindi strings to ISL for possible use in Indian Railways reservation counter. (Kar et al.(2007)Kar, Reddy, Mukherjee, and Raina) It depends on HamNoSys. Raghavan et al.(2013)Raghavan, Prasad, Muraleedharan, and Geetha) proposed a notation based approach for English to ISL translation. It uses a representation for animation and while transforming text to ISL, it queries database. Dasgupta and Basu(2008)) proposed an english text to ISL system. It takes English sentences as input, performs analysis and generates corresponding ISL. In it output is represented in terms of prerecorded video streams. (Dasgupta and Basu(2008))

## 4  System Architecture

The proposed system works by taking Malayalam text as input. The first stage is the POS tagger. Morphological analysis is done to accomplish this. During this step various part of speeches are identified. Then this is fed to optimizer, which removes unwanted words. Output of this module is then given to stemming module, which finds the root word. This is then fed to the animation module. Animation module animates the 3D human character based on the input recieved from the stemming module. Following figure illustrates the whole process.

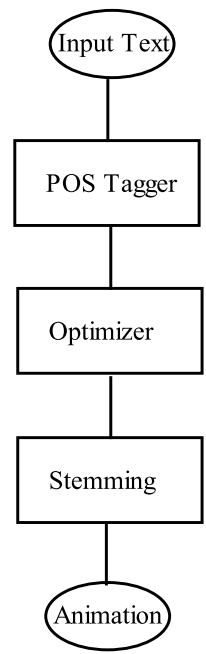

Figure 1: Design

While converting text to ISL, all the input are not directly translated to corresponding sign language. Sign language omits certain words. These task is accomplished with the help of the optimizer module. The words in input text needs to be transformed to the root form before converting it to sign language. Stemming module does this job. After this step we will be getting a sequence of words ready to be converted into sign language. Output from this step is fed to the animation module. For this a properly rigged human character is used. A rule based approach is used to map output generated by stemming module to the animation module.

## 5  Results

For visualisation, a 3 dimensional (3D) computer generated model is used. Animation is done in Free Software tools running on Debian GNU Linux. Animation is modelled as a sequence of key frames. Based on the user input different key frames are nicely blend to produce the corresponding animation. As of now, except complex facial expressions, other features are modelled. Unlike systems that use an intermediate notation to denote sign language, we are directly converting the input to sign language. This reduces the complexity of converting between different formats.

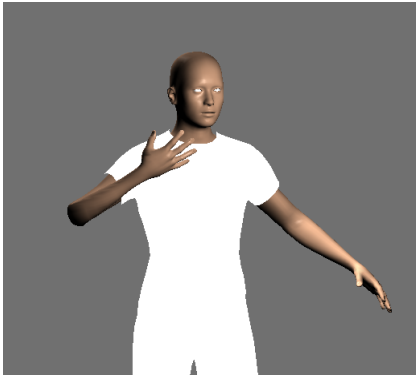

Figure 2: Animation side view

The animation can also be exported to run on smart phones.

# 6 Conclusion and Future Work

This paper presents a prototype Malayalam text to sign language translation system. This paper stresses the need to develop a standardised system for sign language learners in Kerala. The system proposed in this paper can be utilised as a system for standardising sign language use in Kerala. This can easily be ported to mobile handsets and tablets. This system can be also extended to convert text from web sites to sign language form.

As one of the first attempt in Malayalam text to ISL conversion, we hope this work will lead to more developments in social and technological front and thus will result in the development of a standardised system for sign language learners.